\documentclass[journal]{IEEEtran}
\usepackage{mathrsfs}
\usepackage{color}
\usepackage{tikz}
\usepackage{enumerate}
\usepackage{url} 
\usepackage{times}
\usepackage{amsmath}
\usepackage{booktabs}
\usepackage{diagbox}
\usepackage{times}
\usepackage{epsfig}
\usepackage{subfigure}
\usepackage{amssymb}
\usepackage{array}
\usepackage{helvet}
\usepackage{courier}
\usepackage{enumitem}
\usepackage[ruled,linesnumbered]{algorithm2e}
\usepackage{times}
\usepackage{epsfig}
\usepackage{graphicx}
\usepackage{amsmath}
\usepackage{amssymb}
\usepackage{multicol}
\usepackage{multirow}
\usepackage{bbding}
\usepackage{cite}
\usepackage{pgfplots}
\usepackage[switch]{lineno}
\usepackage[pagebackref=true,breaklinks=true,citecolor=blue,linkcolor=blue,letterpaper=true,colorlinks,bookmarks=false]{hyperref}
\usepackage[pagebackref=true,breaklinks=true,letterpaper=true,colorlinks,bookmarks=false]{hyperref}
\usepackage[explicit]{titlesec}
 
\titlespacing{\subsection}{0pt}{1.2ex plus .0ex minus .0ex}{.3ex plus .0ex}
\ifCLASSINFOpdf
\else
\fi
\begin{document}

%

\title{
\begin{small}
Copyright @ 2021 IEEE. Personal use of this material is permitted. However, permission to use this material for any other purposes must be obtained from the IEEE by sending an email to pubs-permissions@ieee.org \\
\end{small}
Human-centric Spatio-Temporal Video Grounding With Visual Transformers}
%
%
%

\author{
  Zongheng Tang, Yue Liao, Si Liu*, Guanbin Li, Xiaojie Jin, Hongxu Jiang, Qian Yu, Dong Xu,~\IEEEmembership{Fellow,~IEEE}
\IEEEcompsocitemizethanks{\IEEEcompsocthanksitem  	 
Zongheng Tang, Yue Liao, Si Liu, Qian Yu, and Hongxu Jiang are with Beihang University, Beijing, China.
Guanbin Li is with the School of Computer Science and Engineering, Sun Yat-sen University, China, and Pazhou Lab, Guangzhou, 510330, China.
Xiaojie Jin is with ByteDance AI Lab, Beijing, China.
Dong Xu is with the School of Electrical and Information Engineering, the University of Sydney, Sydney, Australia.
The Corresponding author is Si Liu (Email: liusi@buaa.edu.cn).
}
}

%
%

\markboth{IEEE TRANSACTIONS ON CIRCUITS AND SYSTEMS FOR VIDEO TECHNOLOGY}%
{Shell \MakeLowercase{\textit{et al.}}: Bare Demo of IEEEtran.cls for IEEE Journals}
%



\maketitle

\begin{abstract}
In this work, we introduce a novel task – Human-centric Spatio-Temporal Video Grounding (HC-STVG). Unlike the existing referring expression tasks in images or videos, by focusing on humans, HC-STVG aims to localize a spatio-temporal tube of the target person from an untrimmed video based on a given textural description. This task is useful, especially for healthcare and security related applications, where the surveillance videos can be extremely long but only a specific person during a specific period is concerned. HC-STVG is a video grounding task that requires both spatial (where) and temporal (when) localization. Unfortunately, the existing grounding methods cannot handle this task well.
We tackle this task by proposing an effective baseline method named Spatio-Temporal Grounding with Visual Transformers (STGVT), which utilizes Visual Transformers to extract cross-modal representations for video-sentence matching and temporal localization.
To facilitate this task, we also contribute an HC-STVG dataset\footnote{The new dataset is available at https://github.com/tzhhhh123/HC-STVG.} consisting of 5,660 video-sentence pairs on complex multi-person scenes. Specifically, each video lasts for 20 seconds, pairing with a natural query sentence with an average of 17.25 words. Extensive experiments are conducted on this dataset, demonstrating that the newly-proposed method outperforms the existing baseline methods.
\end{abstract}


%
\IEEEpeerreviewmaketitle

%
%
%
%

\begin{figure*}[ht]
  \includegraphics[width=0.98\linewidth]{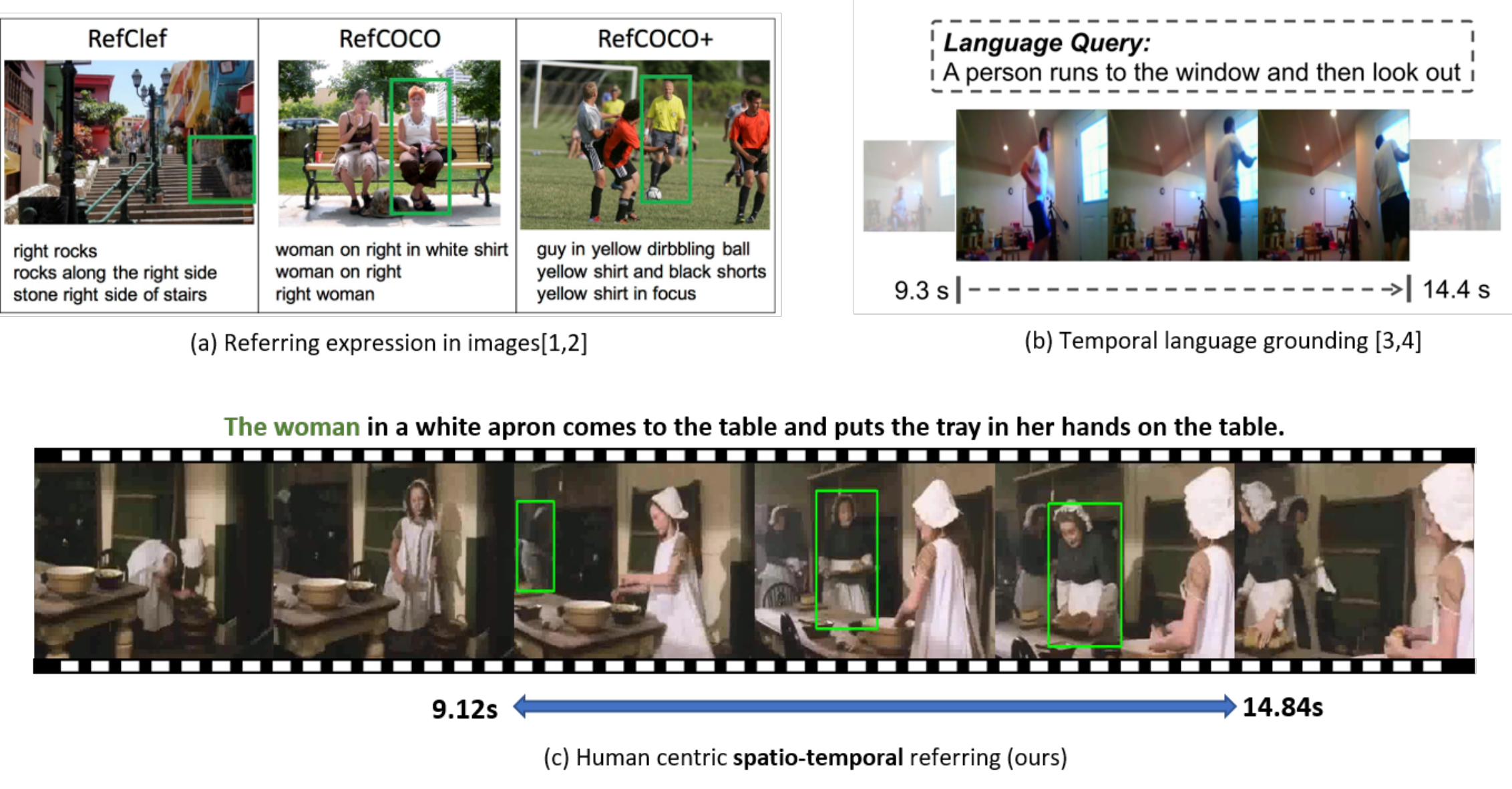}
  \caption{Comparison of different visual grounding tasks. (a) Referring expression in images (e.g., ~\cite{yu2016modeling,kazemzadeh2014referitgame}) is a task of localizing objects/regions in images based on a given query. (b) Temporal video grounding aims at localizing the starting and ending frame corresponding to the described action(s) (e.g.,~\cite{gao2017tall,anne2017localizing}). (c) Human-centric Spatio-temporal Video Grounding is our newly proposed task, which outputs both spatial and temporal localization. The example is from our newly collected HC-STVG dataset. Each example includes a video clip and a corresponding description (top), spatial annotation (i.e., the green boxes), and temporal annotations (i.e., the blue bottom line).  
  }
  \label{fig:teaser}
\end{figure*}

\section{Introduction}
\IEEEPARstart{G}{iven} natural language queries, a visual grounding task aims to localize objects or regions in images or videos, which is an important task in the vision-language research field. It was originated by localizing objects in an image based on short descriptions or sentences~\cite{yu2016modeling,kazemzadeh2014referitgame} (Fig. \ref{fig:teaser}(a)). With the progress on video understanding, recent efforts have been made on referring expression in videos, such as temporal video grounding. Taking Fig. \ref{fig:teaser}(b) for an example, this video grounding task aims to localize a video segment corresponding to the given language query.


Several existing tasks are relevant to the newly proposed HC-STVG task from the task perspective, but they have a different focus.  Specifically,
\textbf{Referring Expression} aims to spatially localize the target object in an image and outputs a bounding box. 
\textbf{Temporal video grounding} aims to determine the temporal boundaries of the target event in a video and thus returns a set of consecutive frames. 
Compared with these two tasks, the HC-STVG task aims to localize the spatio-temporal tube of the target person described in the textual query, which is more challenging as it requires localization both spatially and temporally.
For example, the target person may appear in the video for a long time, but the interval of the event matching the description may be short. It is necessary to utilize both the textual cues (such as actions and relations with other objects) indicated in the query sentence and the visual cues shown in the video to determine the temporal boundaries.
Furthermore, compared to the \textbf{STVG} task, we only consider humans as our referent. This is out of two considerations. First, humans are specific and always the focus of all species in the videos. 
Second, humans produce a wealth of action information and interact with other people or things. Therefore, the HC-STVG task requires a model to do fine-grained reasoning, which is more challenging but have more practical applications in the real world.

Grounding in both space and time domain is often required in real-world scenarios, especially for healthcare and security related applications, where the surveillance videos can be extremely long, but people only care about a specific person and his/her specific behaviors. For example, in a 24-hours monitoring video, a healthcare doctor at the nursing home may want to see how a particular older person was having his/her lunch, or a police officer may need to search for a suspect from the crowded people and locate the process when the suspect conducts a crime. In both cases, when the description is provided, the expected outputs are a sequence of bounding boxes related to the target person in the consecutive frames corresponding to the described action, i.e., a spatio-temporal tube. Unfortunately, such a video grounding setting has not been explored yet. Hence the existing methods cannot well handle this task.


In this work, we introduce a new task named Human-centric Spatio-Temporal Video Grounding (HC-STVG), which aims to localize a spatio-temporal tube of the target person in an untrimmed video given a query description. As shown in Fig. \ref{fig:teaser}(c), based on the description ‘The woman in a white apron comes to the table and puts the tray in her hands on the table’, an HC-STVG model needs to answer: which person is the target and which frames correspond to the described actions.  To the best of our knowledge, it is the first video grounding task centered on humans. There are three key challenges in our HC-STVG task: 
1) It requires grounding referring expressions both spatially and temporally at the same time. 
2) Multi-modality information, such as visual/textural attributes and actions, is often required for localizing a specific person, especially in a complex multi-person scene.
3) It can be difficult to determine the starting and ending frame of an action due to its dynamic nature. This task is meaningful as it involves cross-modality modeling and fine-grained reasoning.


To effectively deal with this task, we propose a new baseline method, termed Spatio-Temporal Grounding with Visual Transformers (STGVT), in which we jointly exploit multi-modality information from videos and textural descriptions by using a visual Transformer. Specifically, our STGVT method consists of four steps: After generating tube proposals from a video by linking bounding boxes in consecutive frames, our method learns the cross-modality interaction between the tube proposals and the query sentence via a Visual Transformer, which is then followed by predicting the matching tube and trimming the irrelevant frames. Although there are many vision-language datasets for the visual grounding tasks, none can support this new task. To address the issue, we contribute an HC-STVG dataset. Precisely, it consists of 5,660 video-query pairs, where 57.2\% of scenes have more than three people. Each query is a sentence with an average of 17.25 words, consisting of rich expressions related to actions and human-object interaction. Based on this new benchmark dataset, we demonstrate the effectiveness of our baseline method for the newly proposed HC-STVG task.  


To summary up, the contributions of this work are three-fold: 
1) We introduce a novel and challenging task, i.e., Human-centric Spatio-Temporal Video Grounding (HC-STVG), which for the first time, focuses on humans in spatio-temporal video grounding. 
2) We build the first human-centric video-description dataset, which can be used as the benchmark dataset for the new task and inspires the subsequent research.
3) We propose an effective baseline method for the new task and demonstrate the effectiveness of the proposed method through extensive experiments.

\begin{figure*}[h]
  \includegraphics[width=0.98\linewidth]{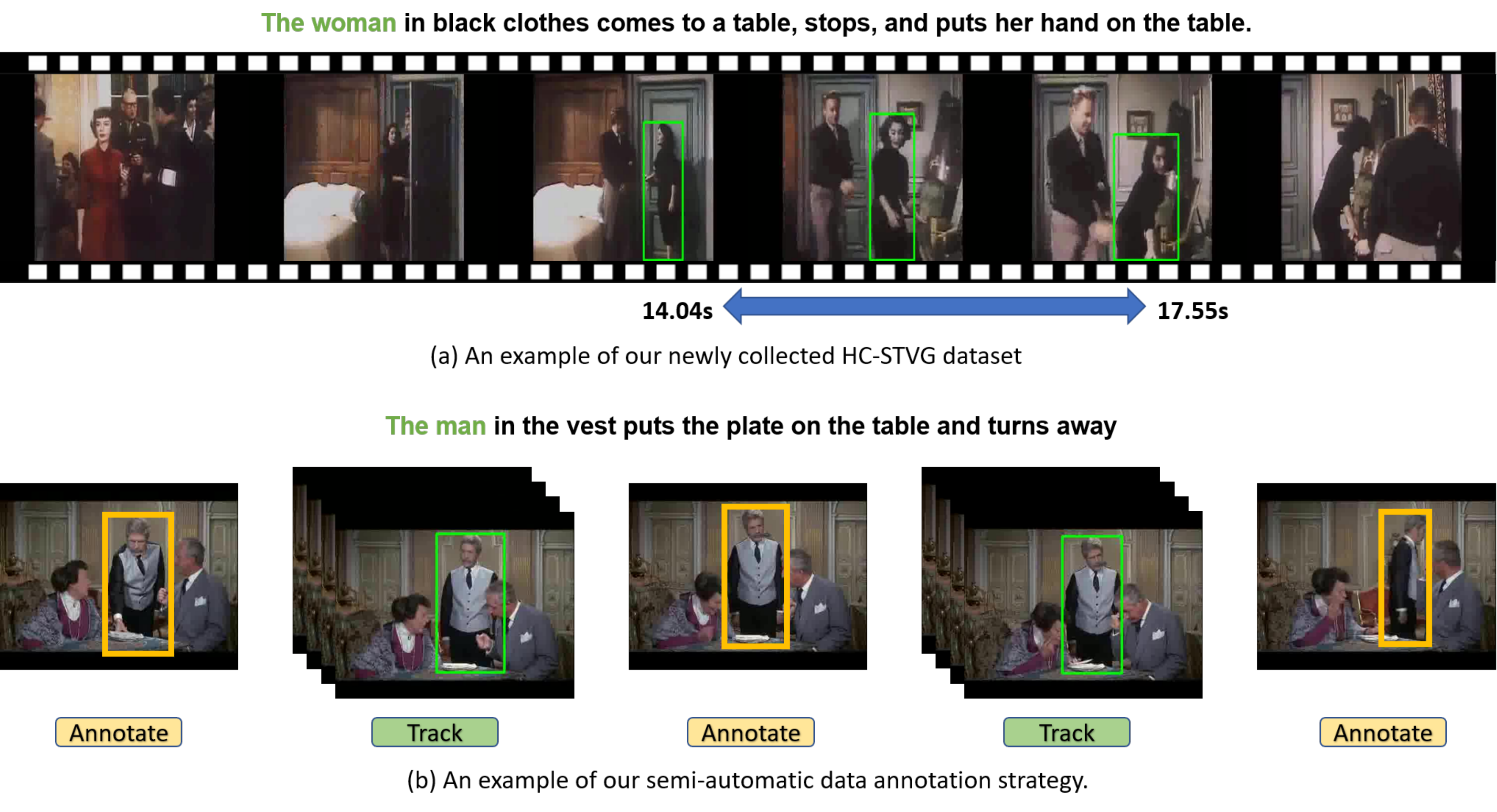}
  \caption{Examples of our newly collected HC-STVG dataset. (a) For a video clip, we collect its text description (top), bounding boxes (green boxes), and temporal annotation (bottom). (b) An example of our semi-automatic annotation strategy. For each video clip, the volunteers are asked to annotate the yellow bounding boxes in three selected keyframes, while the green bounding boxes in the remaining frames are tracked by the method SiamPRN ~\cite{li2018high}.
  }
  \label{fig:track}
\end{figure*}

\section{RELATED WORK}

\subsection{Referring Expression in Images/Videos}
Referring Expression aims to localize the visual object described by any natural language expression ~\cite{yu2016modeling,kazemzadeh2014referitgame,mao2016generation,huang2020ordnet,huang2020referring,hui2020linguistic,Liao_2020_CVPR,yu2018mattnet,liao2019real}. The previous works~\cite{mao2016generation} and~\cite{yu2018mattnet} used a pretrained object detection network or an unsupervised method to generate object proposals and then match these regions with the textual description to select the most relevant proposal. 
To better capture multi-modality context information, MAttNet~\cite{yu2018mattnet} proposed to decompose the referring expression into \textit{subject}, \textit{location}, and \textit{relation}. Moreover, cross-modality correlation filtering is adopted for the referring expression task in real-time~\cite{liao2019real}.
As for referring expression in videos, the WSSTG~\cite{chen2019weakly} is proposed to localize the spatial-temporal tube of an object in a \textit{trimmed} video where the target objects exist through the whole clip. 

\subsection{Visual Transformer}
In recent years, Visual Transformer methods ~\cite{alberti2019fusion,chen2019uniter,li2020unicoder,li2019visualbert,su2019vl,lu2019vilbert,tan2019lxmert} have achieved impressive performance in many Vision-Language tasks, including the Referring Expression tasks.
These works usually employed a transformer-structure model to extract multi-modal feature representation from image-text or video-text pairs and then predict the regions of interest(ROI) based on the cross-modal features.
The existing transformer-based methods proposed for vision-language tasks can be divided into two categories, the single-stream models and the two-stream models. The single-stream models ~\cite{alberti2019fusion,chen2019uniter,li2020unicoder,li2019visualbert,su2019vl} employ a single transformer to learn cross-modal features from the visual and the textual modalities. In contrast, the two-stream models ~\cite{lu2019vilbert,tan2019lxmert} use different encoders to encode information of different modalities while using a co-attention module to learn cross-modal feature representation.

\subsection{Temporal Video Grounding}
The goal of temporal video grounding is localizing the most relevant video segment given a query sentence. The previous works ~\cite{gao2017tall,anne2017localizing,liu2018attentive,liu2018cross,xu2019multilevel} used a temporal sliding-window-based approach over the video frames to generate temporal candidates and choose the most relevant one as the output. Chen et al. ~\cite{chen2018temporally} proposed aggregating the fine-grained frame-by-word interaction between videos and queries. 
Xu et al. ~\cite{xu2019multilevel} and Chen et al. ~\cite{chen2019semantic} proposed to generate query-specific proposals as candidate segments by directly integrating sentence information with each fine-grained video clip.
Ge et al.~\cite{ge2019mac} explored activity concepts in both videos and queries for temporal localization. Zhang et al. ~\cite{zhang2019man} iteratively adjusted the structured graph to deal with the semantic misalignment problem.
Yuan et al.~\cite{yuan2019sentence} utilized the Graph Convolutional Network ~\cite{kipf2016semi} to model the relations between different video clips and proposed a temporal conditioned pointer network to screen the answer.
Unlike these previous video grounding works, which only deal with temporal localization, the newly proposed HC-STVG tackles spatial localization and temporal localization simultaneously.

\section{HC-STVG BENCHMARK}\label{sec:Dataset detail}
The newly proposed HC-STVG task aims to localize the target person spatio-temporally in an untrimmed video. For this task, we collect a new benchmark dataset with spatio-temporal annotations related to the target persons in complex multi-person scenes, together with full interaction and rich action information.



\subsection{Task Formulation}\
In this work, the HC-STVG task is defined as follows.
Given an untrimmed video  $v\,\epsilon\,V$ and a natural language description $s\,\epsilon\,S$, which describes a sequence of actions related to a specific person in $v$, we aim to generate the spatio-temporal tube $T$ (i.e., a sequence of bounding boxes) corresponding to the target person. Specifically, the system needs to locate the segment by determining the starting and ending frames ($l$, $r$) and also generates the bounding boxes of the target person in the located segment.


\subsection{Overview of HC-STVG dataset}
There are 5,660 video-sentence pairs in our newly collected HC-STVG dataset, including 4,500 training pairs and 1,160 testing pairs. 
The average temporal duration of the ground-truth tubes is 5.37 seconds while each sentence has an average of 17.25 words. The duration of videos is further normalized to 20 seconds. 
We ensure that the test samples and the training samples do not originate from the same raw video. 

The key characteristics of our HC-STVG are summarized as follows: (1) human-centric. The dataset contains precise spatio-temporal annotations  and textual description for the person of interest in each video. (2) All videos are captured in complex multi-person scenes, along which 57.2\% of the videos have more than three people while the rest videos have two persons. (3) All sentences have rich descriptions related to the interaction between human-human or human-objects and 56.1\% of the descriptions include both types of interaction.

\begin{figure*}[ht]
	\begin{center}
		\includegraphics[width=0.96\linewidth]{{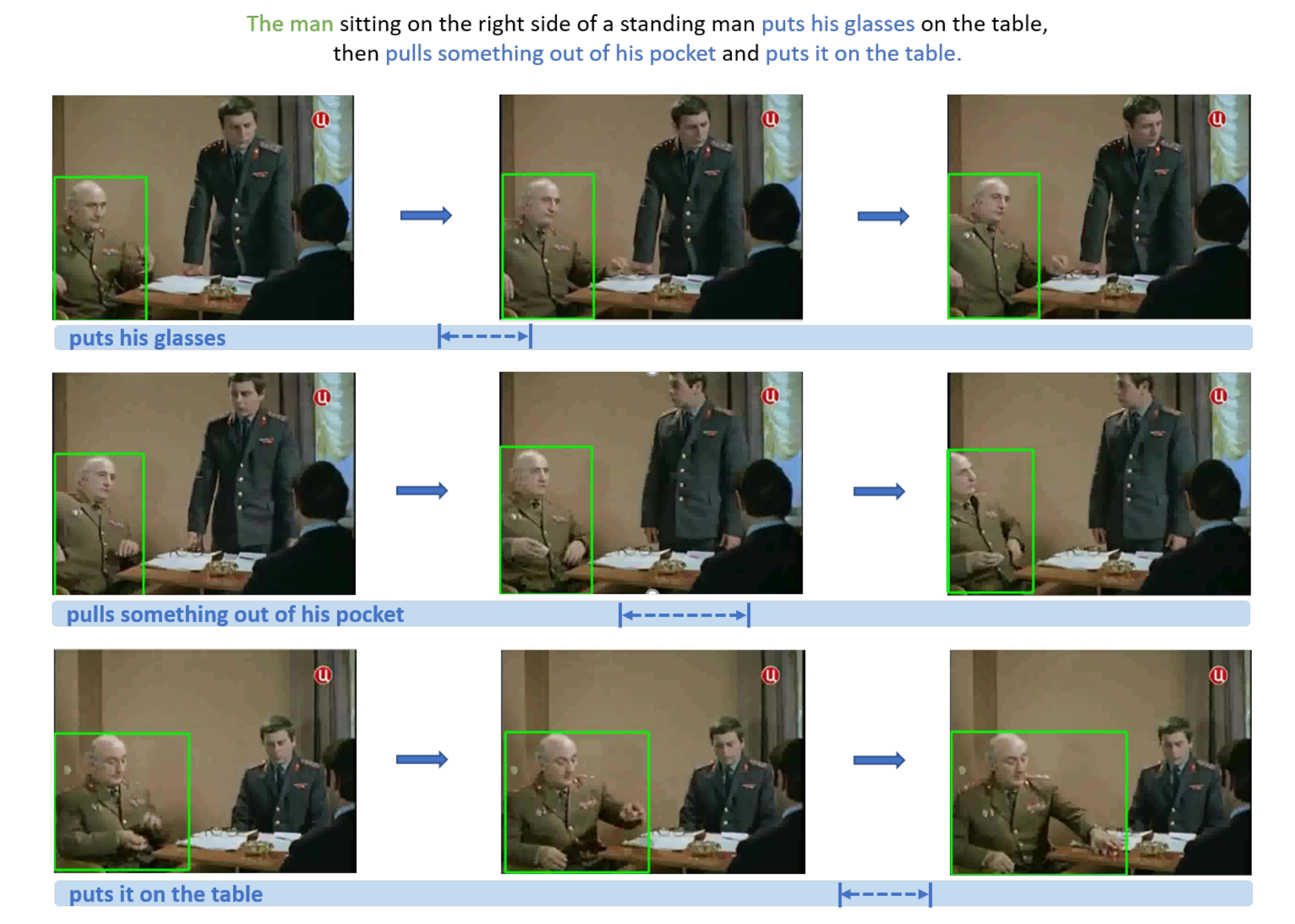}}
	\end{center}
	\caption{An example video-sentence pair in our newly collected dataset. In the example video clip, the target person performs a series of actions. Correspondingly, these actions are described in the sentence, including `puts his glasses on the table',`pulls something out of his pocket', and `puts it on the table'.}
	\label{fig:verb_vis}
\end{figure*}

Figure~\ref{fig:track}(a) and Fig.~\ref{fig:verb_vis} show several examples of our newly collected dataset. As demonstrated in these two figures, the videos in the new dataset are captured in multi-person scenes. Besides, the descriptions have rich expressions related to actions and relations.

\subsection{Dataset Construction}\label{subsec:dataset_construction}
The whole process includes five modules: Raw Video Preparation, Video Span Selection, Video Description, Bounding box annotation, and Video Span Extension. We explain these parts in detail.

{\bfseries Raw Video Preparation}
We use raw video filtered by AVA dataset~\cite{DBLP:conf/cvpr/GuSRVPLVTRSSM18}.
These original videos have been collected on Youtube and have been carefully selected to ensure the diversity and high-quality of the videos. These videos contain a lot of realistic scenes, character actions, and character interactions. Therefore, we can produce more suitable video-description pairs from the videos.

{\bfseries Video Span Selection} 
Annotators watch untrimmed videos sequentially and follow several rules to mark suitable describable spans. To be specific, the chosen spans must be multi-person scenes, thus increasing the difficulty of grounding in our dataset. 
Shot switching is not allowed in the annotated clips because shot switching splits the video content into discrete paragraphs, bringing semantic incoherence. And We limit the shortest video span time to 3s to avoid annotators only tagging the instantaneous actions of the characters instead of continuous action sequences.
Under these restrictions, annotators annotate the precise temporal segment of the target person's actions in the scenario.

{\bfseries Video Description}
The annotator who decides the specific video span is required to write down a description of the target person. 
To collect more diversified and natural expressions, we do not provide any templates and predefined verb lists for annotators to follow. But we still have some guidelines for annotators to make an appropriate textual annotation. 
(1) The description can spatio-temporally locate the target person uniquely, which means that there can not exist another person that satisfies the description and the actions of the target person are traceable. 
(2) We encourage annotators to describe in terms of explicit visual attributes, human actions, and relations, as our goal is to contain a variety of reasoning information. And descriptions of other persons' attributes in the sentence are also encouraged to increase diversity.
(3) As for actions descriptions, in addition to using some transitive verbs (e.g.\ walk, turn) for target persons, workers are encouraged to write human-to-human and human-to-object interactions (e.g.\ hug, follow) as much as possible, which urges the algorithm to pay close attention to the identification and alignment of dynamic visual relation. 
(4) The annotators are encouraged to label successive events that consist of the target character and related actions, which increases the number of verbs in the sentence and differs HC-STVG from the mere action localization task.
(5) Since we only deal with the video images, the annotators are required to focus on the visual content rather than the sound content. At last, after the annotations, we manually review the descriptions to ensure label quality.

{\bfseries Bounding Box Annotation}
Frame-level bounding box annotation is quite crucial for visual grounding tasks to assist in the judgment of location correctness. However, manually labeling bounding boxes for each frame in the video may lead to enormous labor pressure.
To trade off this problem, we adopt the way of manually labeling keyframes and automatically labeling other frames with the tracking algorithm. 
Concretely, we use SiamRPN~\cite{li2018high} to track the target person. We manually annotate three keyframes of a video span (i.e., starting, middle, ending frame) and track the other frames, which effectively outperforms using only one keyframe. We track the person in the annotated temporal clip from front to back and from back to front using keyframes, then we average these two tracking results as an annotation. 
We conduct an additional manual review to correct the tube for samples with apparent differences in two direction tracking results.  

{\bfseries Video Span Extension}
After the video spans are collected and the spatial-temporal annotations are complete, we extend the video clips forward and backward on the temporal ground truth randomly to achieve a fixed total length of time for final query videos, and the extended time also serves as a negative sample for temporal localization.
As localizing descriptions in query videos may be ambiguous after time extension, in HC-STVG, additional annotators review the total video span to guarantee there is no unambiguous referring. 

\begin{table}[ht]
\centering
\caption{Statistics of Different Datasets. \\ ALS stands for average length of sentences. BA and TA stand for Bounding Box annotation and  Temporal annotation, respectively. HC stands for Human-centric.}
\begin{tabular}{p{2.5cm}p{1cm}p{1cm}p{0.5cm}p{0.3cm}p{0.3cm}p{0.3cm}}
\hline 
Dataset & \#Queries & \#Videos & ALS & BA & TA & HC \\
\hline 
DiDeMo~\cite{anne2017localizing} & 40,543 & 10,464   & 8.0           & $\times$              & \checkmark                     & $\times$             \\
CharadesSTA~\cite{gao2017tall} & 16,128 & 6,670  & 7.2  & $\times$              & \checkmark                     & \checkmark               \\
TACoS~\cite{rohrbach2014coherent} & 18,818 & 7,206 & 10.5 & $\times$        & \checkmark   & \checkmark   \\
ActivityNet-C~\cite{krishna2017dense}  & 71,942 & 12,460 & 14.8        & $\times$              & \checkmark                     & \checkmark               \\
VID-sentence~\cite{chen2019weakly}   & 7,654  & 5,318  & 13.2        & \checkmark              & $\times$                   & $\times$             \\
VidSTG~\cite{zhang2020does} &44,808  & 6,924 & 11.12       & \checkmark              & \checkmark                     & $\times$    \\  
\hline 
HC-STVG Dataset & 5,660 & 5,660  & 17.25       & \checkmark              & \checkmark                     & \checkmark    \\  
\hline 
\label{tab:datasets}
\end{tabular}
\end{table}

\subsection{Comparison with the Existing Datasets}
We compare the existing video grounding datasets with our newly collected HC-STVG dataset in Table~\ref{tab:datasets}. The DiDeMo dataset~\cite{anne2017localizing} only provides temporal annotation for the temporal localization task. The TACoS dataset~\cite{DBLP:journals/tacl/RegneriRWTSP13}, ActivityNet Captions dataset~\cite{krishna2017dense}, and Charades-STA dataset~\cite{gao2017tall} are all human-centric datasets as their videos capture human actions (e.g., cooking). Similar to DiDeMo dataset, these datasets are collected for the temporal localization task, and they do not have annotations at the bounding box level. The VID-sentence dataset ~\cite{chen2019weakly} provides the bounding-box-level annotations. However, it only focuses on spatial localization. Video clips in this dataset are all trimmed, hence it is not suitable for temporal localization. Among all datasets, the most relevant dataset is the VidSTG dataset~\cite{zhang2020does}. It is extended from the dataset Vidor~\cite{2019Annotating}, which is a dataset originally collected for detecting relations in videos. The VidSTG dataset provides both bounding-box-level annotations and temporal annotations. However, this dataset focuses on the task of relation referring between subjects and objects. In contrast, our HC-STVG dataset focuses on localizing the target person spatially and temporally from multi-person scenes based on the query sentences. Furthermore, on average the length of each sentence in our dataset is the longest among these datasets, and the descriptions from this dataset contain rich expressions related to actions and human-object interaction. The details of all these datasets are summarized in Table~\ref{tab:datasets}.

\begin{figure*}[ht]
    \includegraphics[width=1.0\linewidth]{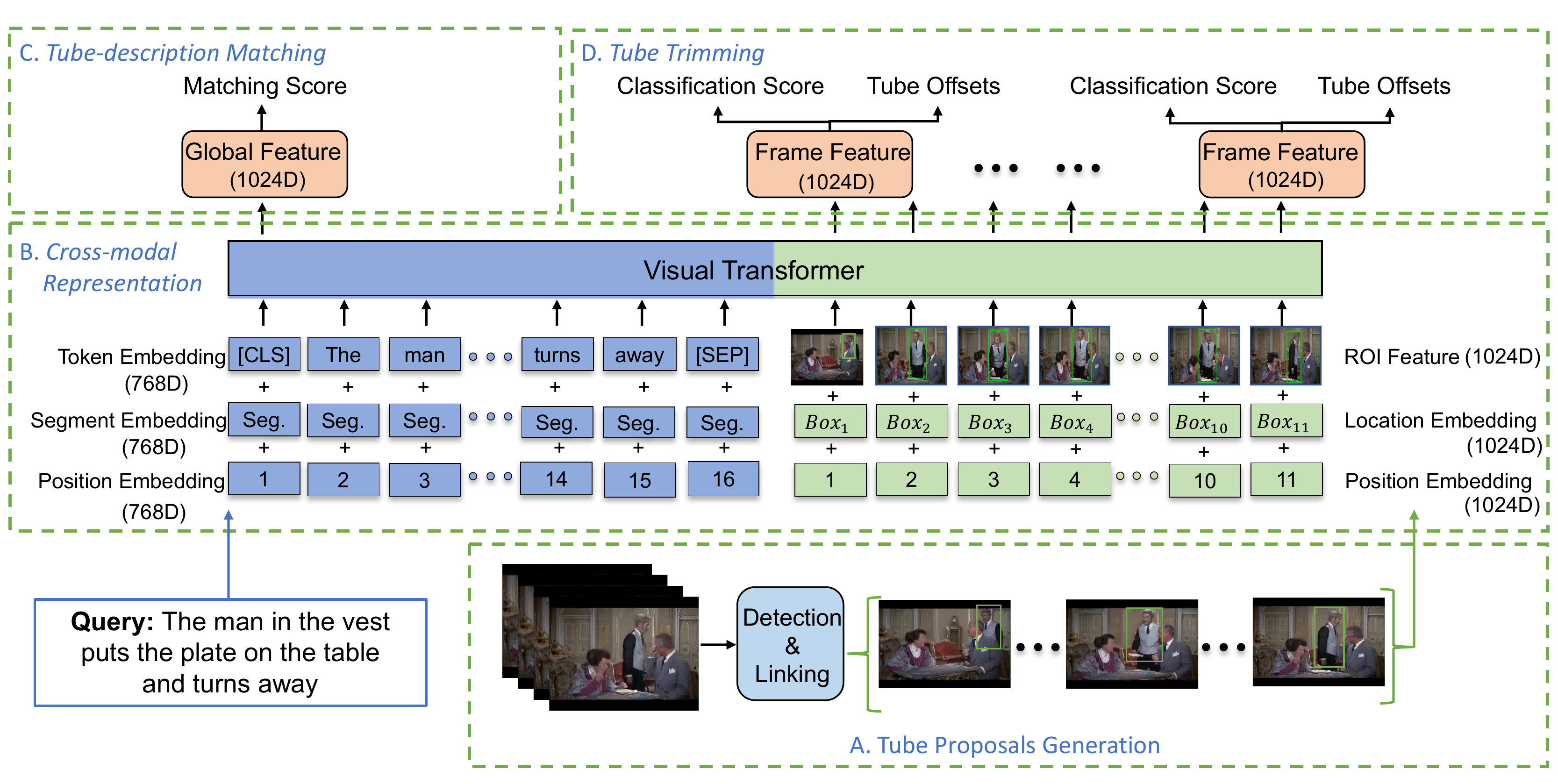}
  \caption{The framework of our STGVT. The framework has four modules, which correspond to four steps explained in Methodology. Specifically, given a query sentence and video, it first generates tube proposals (A), and then each tube-description pair is fed into the visual transformer (B). The output feature of the visual transformer is used for predicting matching score, classification score, tube offsets , which to be used for tube selection (C) and tube trimming (D).}
  \label{fig:framework}
  
\end{figure*}
\section{METHODOLOGY} \label{sec:method}
The overall framework of our STGVT method is shown in Fig.~\ref{fig:framework}. Given a video $v$ and a query sentence $s$ depicting a person performing a series of actions, our STGVT  method aims to output a spatio-temporal tube $T$ corresponding to the description. The method consists of four steps:
\textbf{1)} it first generates the tube proposals (Sec.~\ref{sec:tube}). \textbf{2)} Then a visual transformer takes a pair of the tube proposal and the query sentence as the input, and extracts the cross-modal features, including the global feature and the frame features (Sec.~\ref{sec:cross_rep}). \textbf{3)} A matching classifier is adopted to predict a matching score for each tube-sentence pair. The tube proposal with the highest matching score will be proceeded in the final step (Sec.~\ref{sec:matching}). \textbf{4)} Based on the selected tube, a temporal trimming module is proposed to trim the irrelevant frames and produce the final spatio-temporal localization results. (Sec.~\ref{sec:trimming}).

In this section, we first describe each step of the proposed method and then introduce the loss functions (Sec.~\ref{sec:loss}) and the inference process (Sec.~\ref{sec:inference}). 

\subsection{Tube Proposals Generation}\label{sec:tube}

Given a video, we first use the region proposal network~\cite{ren2015faster} to detect the bounding boxes $B_{t,i}$ in each frame, where $B_{t,i}$ represents the $i$-th bounding box in the $t$-th frame. Basically, we follow the ACT method~\cite{kalogeiton2017action} to link the bounding boxes in consecutive frames to form spatio-temporal tube proposals $T_p$. 
This process is essential to group the bounding boxes containing similar visual content in adjacent frames. 
Different from the method~\cite{kalogeiton2017action}, we compute the similarity score between two bounding boxes (i.e., $B_{t,i}$ and $B_{t+1,j}$) as follow:
\begin{center}
\begin{equation}
\label{eq:link_tube}
\begin{split}
S_(B_{t,i},B_{t+1,j}) =  \lambda_i*IoU(B_{t,i}, B_{t+1,j})  \\ 
+ \lambda_c*D_{cosine}(f(B_{t,i}),f(B_{t+1,j})) \\ 
+ \phi(B_{t,i}) +  \phi(B_{t+1,j}),
\end{split}
\end{equation}
\end{center}
where $f(\cdot)$ is a pretrained feature extractor, $D_{cosine}(\cdot,\cdot)$ is the cosine function, $\phi$ is the confidence score of the bounding boxes, and  $\lambda_i,\lambda_c$ are the weights of the IoU score and the cosine score. In practice, we set them as 0.7 and 0.3 respectively.



\subsection{Cross-modal Representation}
\label{sec:cross_rep}

The next step is to extract cross-modal features for each pair of the tube proposal $T_p$ and the query description $s$. We adopt Visual Transformer in this step due to its great success for the cross-modal tasks~\cite{lu2019vilbert}. The visual transformer takes the visual and textual inputs separately and models their interaction through a set of transformer layers based on the co-attention mechanism. Specifically, following~\cite{devlin2018bert}, we use the position information, token information, and segment embedding extracted from the query description $s$ as the textual inputs; while each temporal position, visual feature, and spatial location from tube proposal $T_p$ are used as the visual inputs for each tube proposal $T_p$. The temporal position refers to the frame index, the visual features are extracted from the bounding box in each frame of $T_p$, and the bounding boxes' coordinates are encoded as the location
embedding. 

The visual transformer will output two cross-modal features $f^{global}$ and $f^{frame}$. When using the textual inputs as the Query and the visual inputs as the Key and Value, $f^{global}_p$ ($p$ indicates the $p$-th tube-sentence pair) is the element-wise product of the visual feature and textual feature, which are extracted from the first token position. When the visual inputs are used as Query and the textual inputs are used as Key and Value, the model generates the feature $f^{frame}_{p,t}$ for each bounding box in each frame of $T_p$ ($p$ and $t$ refer to the $t$-th frame of the $p$-th tube). These two features are used to predict the matching scores of $T_p$ at the tube-level and the frame-level, which to be explained below. 

\subsection{Tube-description Matching}
\label{sec:matching}
In this step, our method will verify if the input tube proposal $T_p$ is matching with the query description $s$. As mentioned above, the visual transformer extracts the global feature $f^{global}_p$ for each pair of $T_p$ and $s$. We feed $f^{global}_p$ into a binary classifier to predict its matching score $M'_p \in [0,1]$. $T_p$ with the highest matching score will be selected and fed into the last step.

When training the classifier, it is worth noting that the positive and negative samples can be imbalanced if we only treat the ground-truth tube, $T_{GT}$, as the positive sample. To tackle this problem, we relax the criteria for the positive samples by introducing two scores, $s_{overlap}$ and $s_{IoU}$. The tube proposals which simultaneously satisfy the following two conditions, $s_{overlap}>=0.9$ and $s_{IoU}>0.5$, are treated as the positive samples. Specifically, (1) $s_{overlap}$ is defined as follow:
\begin{equation}
\label{eq:overlap}
s_{overlap} =  \frac{\left | T_p \cap  T_{GT} \right |}{\left | T_{GT} \right |},
\end{equation}
$\left | T_{GT} \right |$ refers to the total number of  frames in the ground-truth tube. This score reflects the ratio of the total number of the intersected frames between $T_p$ and $T_{GT}$ over all frames in $T_GT$.
2) The average IoU score $s_{IoU}$ is defined as follow:
\begin{equation}
\label{eq:IoU}
s_{IoU} = \frac{1}{\left | T_p \cap  T_{GT} \right |}\sum _{t\in {T_p \cap  T_{GT}}}{IoU \left ({B'_{t}},{B_{t}}\right)},
\end{equation}
where ${B'_t}$ and ${B_t}$ are the detected/ground-truth bounding boxes of $T_p$ and $T_{GT}$ at frame $t$, respectively. A tube is treated as a negative sample when its $s_{IoU}$ is lower than 0.2. 


\subsection{Tube Trimming}
\label{sec:trimming}


Given the dynamic characteristics of actions, the selected tube proposal $T_p$ may contain redundant transition frames. Hence it is required to trim $T_p$ to output the final predicted tube $T$. We compute the relevance score of each frame in $T_p$ to the query $s$, and predict the regression offsets. Our trimming module consists of two subnets, a classification subnet and a boundary regression subnet, which will be detailed below.

\subsubsection{Classification Subnet}
The classification subset is used to predict if a video frame should be kept in the final tube -- the spatio-temporal localization result $T$, or not. As introduced in Sec.~\ref{sec:cross_rep}, the transformer outputs $f^{frame}_{p,t}$ for each frame of $T_p$. The classifier takes $f^{frame}_{p,t}$ as the input to predict a relevance score $C'_{p,t}$ indicating how relevant the $t$-th frame is to the query sentence $s$. 



\subsubsection{Boundary Regression Subnet}
Besides the relevance score $C'_{p,t}$, we also adopt a network to predict the temporal offsets, which reflects how far away the current frame is from the ground-truth temporal boundary. This subset's architecture is the same as the classification subset except that it has two outputs. Specifically, for a positive frame at the $t$-th position, if the ground-truth tube spans from the $l$-th frame to the $r$-th frame (i.e., $R_{GT}=[l,r]$), the regression target is $O_{p,t}=(\delta{l}, \delta{r})$, 
\begin{equation}
\delta{l} = \frac{t - l}{N} \ \ \ \ \ \delta{r} = \frac{r - t}{N}, 
\end{equation}
where $N$ is the number of frames in  tube $T_p$. $\delta{l}$ and $\delta{r}$ refer to the temporal offset from the $t$-th frame to the left and the right boundary, respectively.


\subsection{Loss Functions}
\label{sec:loss}
Given the tube-level matching prediction score $M_p'$, the frame-level relevance prediction score $C'_{p,t}$, and boundary regression prediction result $O'_{p,t}$, and their corresponding ground-truth labels $M_p$, $C_{p,t}$, $O_{p,t}$, the total loss is defined as follows,
\begin{center}
\begin{equation}
\label{eq:loss}
\begin{split}
L = \sum_p \{\lambda_1*L_{match}\left (M_p',M_p \right ) \\ +\lambda_2*\mathbb{I}_{\left \{ M_p=1 \right \}}  \frac{1}{N}  \sum_t{L_{cls}\left (C_{p,t}',C_{p,t} \right )}\\+
\lambda_3*\mathbb{I}_{\left \{ M_p=1,C_{t}=1 \right \}}  \frac{1}{N_{pos}}  \sum_t{L_{reg}\left (O'_{p,t},O_{p,t}\right)}\}
\end{split}
\end{equation}
\end{center}
where $L_{match}$ is the cross-entropy loss for the tube-description matching module, $L_{cls}$ and $L_{reg}$ is a cross-entropy loss and IoU loss (i.e., $-ln(\frac{O'_{p,t} \cap O_{p,t}}{O'_{p,t} \cup O_{p,t}})$) for the classification subset and the boundary regression subset, respectively. $N$ and $N_{pos}$ denote the number of frames and the \textit{positive} frames in the tube $T_p$. 
It is worth noting that, except for the tube proposals generation network, the rest of the proposed modules can be trained in an end-to-end fashion. However, the loss $L_{cls}$ and $L_{reg}$ in Eq.~\ref{eq:loss} are only computed for the positive tubes (i.e., $I_{(M_p=1)}$) and positive frames ($I_{(M_p=1,C_{p,t}=1)}$).

\subsection{Inference}
\label{sec:inference}
During the inference process, given a query sentence $s$ and a video $v$, our proposed method: 
\textbf{(1)} first generates $P$ tube proposals $T_p$ $(p=1,2,...,P)$. Each tube proposal $T_p$ and the query sentence $s$ form a pair, which will be fed into the visual transformer. \textbf{(2)} For each pair of $T_p$ and $s$, the visual transformer extracts cross-modal features, $f^{global}_p$ and $f^{frame}_{p,t}$. \textbf{(3)} The tube-description matching module takes the global feature $f^{global}_p$ as the input and predicts a matching score $M’_p$. The tube with the highest matching score $M’_p$ is selected and further trimmed. 
\textbf{(4)} For all frames in the selected tube $T_p$, the classification subnet and the regression subnet take the frame feature $f^{frame}_{p,t}$ as the input, and output the classification score $C’_{p,t}$ and $O’_{p,t}=(\delta{l}, \delta{r})$ for each frame. 
We follow the method DEBUG~\cite{lu2019debug} to determine the temporal boundary of the selected tube. Specifically, we start from the frame with the highest $C’_{p,t}$. Given its predicted regression offsets $O’_{p,t}=(\delta_l, \delta_r)$, we can produce an initial range $R_{init}=(t-\delta_{l}*N,t + \delta_{r}* N)$. Next, we compute $R_t$ for the remaining $N-1$ frames. If the predicted range of the $t$-th frame, $R_t$, has overlap with $R_{init}$, then $R_t$ is merged with $R_{init}$. This process is repeated for all frames whose $C'_{p,t}$ higher than a pre-defined threshold $\epsilon$, which will produce the final predicted tube.

\section{EXPERIMENTS}
\subsection{Evaluation Metrics  \& Implementation Details}
The main evaluation metric for HC-STVG is \textbf{\textit{vIoU}}, and it is defined as follow:
\begin{equation}
vIoU = \frac{1}{\left | {T_{union}}  \right |}\sum _{t\in {T_{inter}}}{IoU \left ({B'_t},{B_t}\right)},
\end{equation}
where $T_{inter}=T_p \cap  T_{GT} $ refers to the intersected frames between the predicted tube $T_p$ and the ground-truth (GT) tube $T_{GT}$, $T_{union}=T_p \cup  T_{GT}$ represents the union set of the predicted frames and the GT frames, ${B'_t}$ and ${B_t}$ are the predicted bounding box and the ground truth bounding box at frame $t$. \textbf{$vIoU$} reflects the accuracy of the predicted spatio-temporal tubes. The \textbf{\textit{vIoU@perc.}} score stands for the percentage of predicted tubes whose $vIoU$ is larger than \textit{perc.} and $\mathbf{m_{vIoU}}$ is the average of $vIoU$s score over the whole test set.

During the inference process, we 
sample one frame from every six frames of the tube before feeding the sampled frames into the visual transformer in order to reduce redundancy. The maximum number of bounding boxes detected in each image is set as 101. We use Mask R-CNN pretrained on the Visual Genome dataset~\cite{krishna2017visual} to extract the visual features from each bounding box. 
Every query sentence $s$ is truncated or filled to the maximum length of 40 words. The loss weights $\lambda_1$, $\lambda_2$, $\lambda_3$ are empirically set as 1, 1 and 2, respectively. During the training process, we use Adam optimizer~\cite{duchi2011adaptive}. The initial learning rate and batch size are empirically set as 2e-5 and 32, respectively.

\begin{table}[h]
\centering
\caption{Performance (\%) over $m\_vIoU$, $vIoU@0.3$ and $vIoU@0.5$ on the VidSTG Dataset of the declarative sentence.}
\begin{tabular}{|c|ccc|}
\hline
\multirow{2}{*}{Method} & \multicolumn{3}{c|}{Declarative Sentence} \\ 
                        & m\_vIoU     & vIoU@0.3     & vIoU@0.5     \\
\hline
Random                  & 0.69\%      & 0.04\%       & 0.01\%       \\ 
GroundeR~\cite{yu2017joint} + TALL~\cite{gao2017tall}         & 9.78\%      & 11.04\%      & 4.09\%       \\ 
STPR~\cite{yamaguchi2017spatio} +   TALL~\cite{gao2017tall}           & 10.40\%     & 12.38\%      & 4.27\%       \\ 
WSSTG +   TALL~\cite{gao2017tall}          & 11.36\%     & 14.63\%      & 5.91\%       \\ 
GroundeR~\cite{yu2017joint} + L-Net~\cite{chen2019localizing}        & 11.89\%     & 15.32\%      & 5.45\%       \\ 
STPR~\cite{yamaguchi2017spatio} +   L-Net~\cite{chen2019localizing}          & 12.93\%     & 16.27\%      & 5.68\%       \\ 
WSSTG~\cite{chen2019weakly} +   L-Net~\cite{chen2019localizing}         & 14.45\%     & 18.00\%      & 7.89\%       \\ 
STGRN~\cite{zhang2020does}                   & 19.75\%     & 25.77\%      & 14.60\%      \\ 
STGVT                   & \textbf{21.62\%}     & \textbf{29.80\%}      & \textbf{18.94\%}  \\ \hline
\end{tabular}
\label{tb:vid}
\end{table}

\begin{table}[h]
\centering
\caption{Performance (\%) of different methods on the HC-STVG Dataset in term of $m\_vIoU$, $vIoU@0.3$ and $vIoU@0.5$}
\begin{tabular}{p{3.5cm}p{1cm}p{1.25cm}p{1.25cm}}
\hline 
Method & $m\_vIoU$ & $vIoU@0.3$ & $vIoU@0.5$ \\
\hline
Random                   & 0.71\%  & 0.03\%   & 0\%      \\
TALL~\cite{gao2017tall}+WSSTG~\cite{chen2019weakly}                & 13.37\% & 19.95\%  & 7.33\%   \\  
2D-TAN~\cite{zhang2019learning}+WSSTG~\cite{chen2019weakly}              & 15.43\% & 19.83\%  & 6.81\%   \\
\hline
STGVT~($w/o~trimming$)  & 16.93\%  & 21.29\%  & 6.64\%       \\
STGVT~($w/o~pretraining$)       & 17.46\% & 25.09\%  & 7.76\%   \\
STGVT       & \textbf{18.15}\% & \textbf{26.81}\%  & \textbf{9.48}\%  \\
\hline 
\label{tb:results}
\end{tabular}
\end{table}

\subsection{Baseline Methods}
Considering that the existing methods cannot be directly applied for the human-centric spatio-temporal video grounding task, we form two baselines by combining the existing methods on video grounding. Specifically, \textbf{TALL}~\cite{gao2017tall} and \textbf{2D-TAN}~\cite{zhang2019learning} are the two methods for temporal video localization, while \textbf{WSSTG}~\cite{chen2019weakly} method is proposed for localizing spatio-temporal tubes in \textit{trimmed} videos. So we first apply TALL or 2D-TAN to produce a temporary segment from each \textit{untrimmed} video based on each given sentence, and then utilize the WSSTG method to localize spatio-temporal tubes in the generated segment. The \textbf{Random} method localizes the temporal segment and spatial regions randomly. 
It is worth noting that we replaced the tube-description matching module with the WSSTG method, while keeping the entire training process fully supervised.
For a fair comparison, we the same strategy for all these methods to generate the tube proposals, as introduced in Section IV-A.

\subsection{Experimental Results}
We report the results of our STGVT method and the baseline methods in Table \ref{tb:results}. 
One can observe that our STGVT method outperforms all baseline methods. Our method achieves the $m_{vIou}$ score of  18.15\%, which outperforms baseline methods by a significant margin. Our method also surpasses the baselines in terms of $vIoU@0.3$ and $vIoU@0.5$.

To further verify the effectiveness of our method, we also conducted experiments on VidSTG~\cite{zhang2020does} dataset. VidSTG is a suitable dataset for Spatio-Temporal Video Grounding task, but it is not human-centric. Thus we considered all the categories in the library when generating tubes, and use the same method to select the target tube and trim it. Experimental results in  Table~\ref{tb:vid}.  show that our method is also effective on different databases.

\subsection{Ablation Study}
\subsubsection{Effectiveness of Tube-description Matching Module}
In our proposed method, the tube-description matching module outputs a raw prediction of the spatio-temporal tube, which will further be trimmed by the tube trimming module. We evaluate the quality of the raw predictions by comparing our method with the WSSTG~\cite{chen2019weakly} method in Table~\ref{tb:match}. It is worth noting that (1) we directly use the tube without any temporal trimming as the final prediction, (2) both ours and WSSTG use the same method for generating tube proposals.

\begin{table}[h]
\centering
\caption{Tube matching accuracy results on the HC-STVG dataset}
\begin{tabular}{p{2cm}p{1.5cm}p{1.5cm}p{1.5cm}}
\hline
Method & $m\_vIoU$ & $vIoU@0.3$ & $vIoU@0.5$ \\
\hline
Random  & 2.40\%  &  1.72\%   & 0.44\%       \\
WSSTG~\cite{chen2019weakly}   & 12.96\%  & 16.23\%  & 4.35\%       \\
STGVT   & \textbf{16.93}\%  & \textbf{21.29}\%  & \textbf{6.64}\%       \\
\hline 
\label{tb:match}
\end{tabular}
\end{table}

\subsubsection{Effectiveness of Tube Trimming Module}
In this section, we demonstrate the effectiveness of the tube trimming module in our proposed method. As shown in Table~\ref{tb:results}, our full method (\textit{STGVT}) significantly outperforms the simplified version without the tube trimming module (termed as \textit{STGVT~(w/o~trimming)}). The performance improvement of our \textit{STGVT} over \textit{STGVT~(w/o~trimming)}  indicates the effectiveness of the tube trimming module. However, it is worth noting that, even without using the tube trimming module, our simplified version \textit{STGVT~(w/o~trimming)} method still achieves better performance compared with other baseline methods. 
Besides, we compared the results of the trimmed positive tube, as shown in Table~\ref{tb:tem_results}, the result of tube selection we chose significantly outperforms the baseline method.

\begin{table}[h]
\centering
\caption{Temporal grounding Performance (\%) over $m\_tIoU$ and $m\_vIoU$}
\begin{tabular}{p{3.5cm}p{1.5cm}p{1.5cm}}
\hline 
Method & $m\_vIoU$ & $m\_tIoU$ \\
\hline 
Random                    & 0.71\%  & 10.50\%    \\
TALL~\cite{gao2017tall} +$Tube_{pos}$         & 31.64\% & 44.31\%   \\   
2D-TAN~\cite{zhang2019learning}+$Tube_{pos}$       & 32.89\% & 46.20\%    \\   
STGVT+$Tube_{pos}$        & \textbf{34.71}\% & \textbf{48.64}\%  \\
\hline 
\label{tb:tem_results}
\end{tabular}
\end{table}




\begin{figure*}[h!]
    \begin{center}
    	\includegraphics[width=1.0\linewidth]{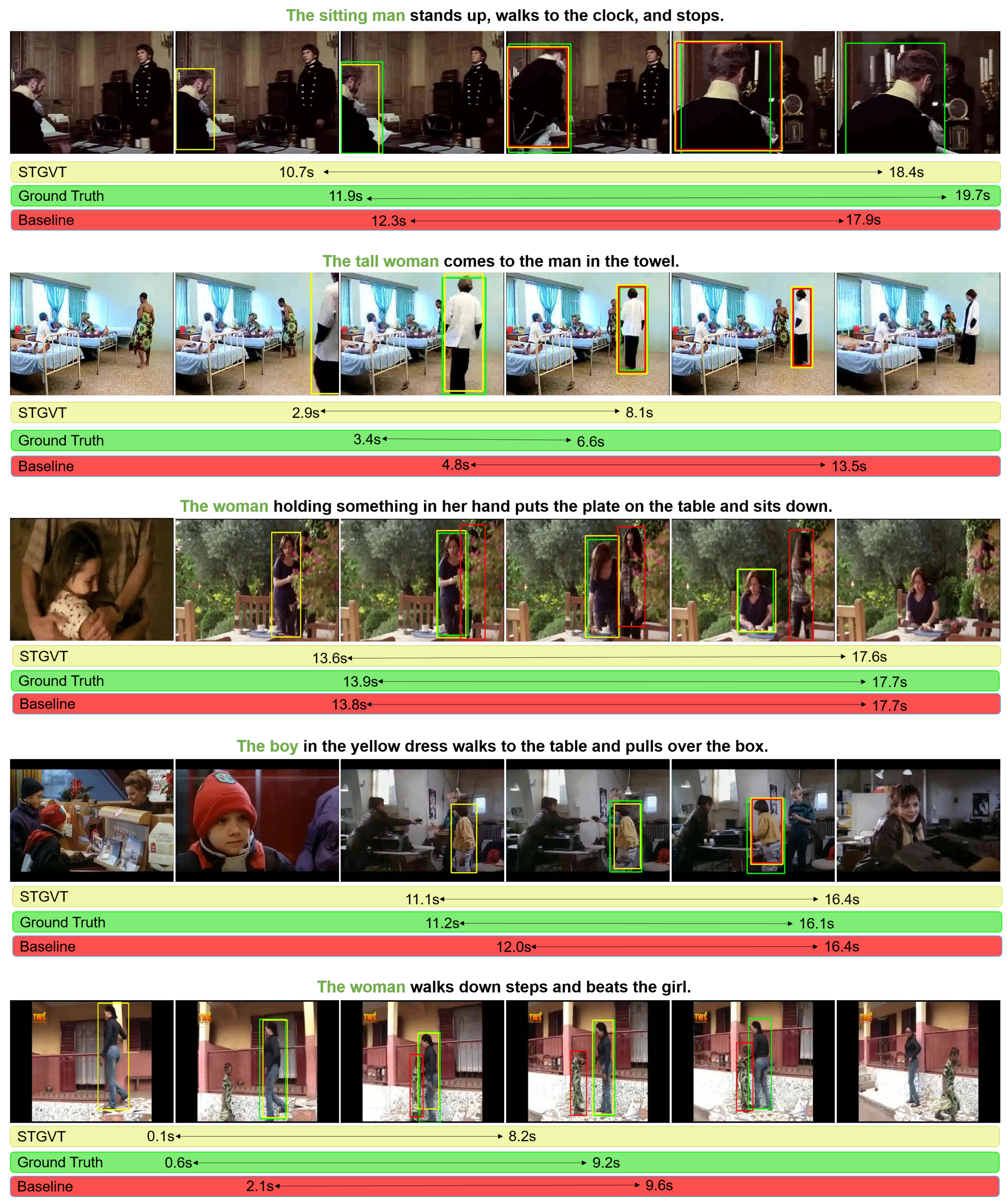}
    \end{center}
    \centering
  \caption{
Representative results produced by our method and the baseline method (WSSTG+TALL). In the 2nd and the 5th examples, our method achieves better temporal localization results when compared with the baseline method. While in the 1st, the 3rd, and the 5th examples, our method achieves more accurate spatial localization results. }
  \label{fig:res}
\end{figure*}

\subsubsection{Effectiveness of the Pretraining Strategy}
Since we adopt the visual transformer \textit{pretrained} based on the Conceptual Captions dataset~\cite{sharma2018conceptual}, we wonder whether the performance improvement is due to the pretraining strategy. So we perform another experiment by training the visual transformer from scratch\footnote{The visual transformer has two branches, one for textual feature while the other for visual feature. All methods except for the baseline method `Random' use a pretrained transformer for the textual branch.} (and our method is referred as \textit{STGVT~(w/o~pretraining)}.
As shown in Table~\ref{tb:results}, 
we observe that: 1) our method \textit{STGVT~(w/o~pretraining)} outperforms existing baseline methods; 2) using the pretraining strategy can improve the performance.

\subsection{Qualitative Analysis}
We provide some visualization results in Fig.~\ref{fig:res}.
These examples show that our method outperforms the baseline method for video grounding both spatially and temporally. Specifically, in most examples (except the 3rd one), there are more overlap between our temporal tube prediction results (see the yellow line) and the ground-truth (see the green line) when compared with the baseline method TALL~\cite{gao2017tall} + WSSTG~\cite{chen2019weakly}(see the red line), which indicates our method can localize the video segment more precisely. In the 3rd example, both our method and the baseline method achieve good performance in temporal grounding, but our method generates more accurate bounding boxes. 

\section{Conclusion}
In this work, we have introduced a novel task Human-centric Spatio-Temporal Video Grounding (HC-STVG). Furthermore, we have contributed a new HC-STVG dataset with rich spatio-temporal tube annotations for videos and the corresponding descriptive sentences.
We have also proposed a baseline method named STGVT, which takes advantage of the Visual Transformers to tackle the HC-STVG task. Comprehensive experiments have demonstrated that our method outperforms the existing grounding methods for the HC-STVG task.
\\

\noindent{\bfseries Acknowledgement:}
This research is supported in part by the National
Key Research and Development Project of China
(No. 2018AAA0101900), the National Natural Science Foundation of China (Grant
61876177), Beijing Natural Science
Foundation (4202034), the Guangdong Basic and Applied Basic Research Foundation (Grant No.2020B1515020048), Fundamental Research Funds
for the Central Universities, Zhejiang Lab (No. 2019KD0AB04).


%





\ifCLASSOPTIONcaptionsoff
  \newpage
\fi



%



{
\bibliographystyle{IEEEtran}
\bibliography{bare_jrnl}
}

\end{document}